\documentclass[11pt]{article}

\usepackage[preprint]{acl}
\usepackage{times}
\usepackage{latexsym}
\usepackage{amsmath,amssymb}
\usepackage{makecell}
\usepackage{multirow}

\usepackage[T1]{fontenc}
\usepackage[utf8]{inputenc}

\usepackage{microtype}

\usepackage{inconsolata}

\usepackage{graphicx}
\usepackage{subcaption}
\usepackage{caption}

\title{MaBERT: A Padding-Safe Interleaved Transformer–Mamba Hybrid Encoder for Efficient Extended-Context Masked Language Modeling}

\author{
  Jinwoong Kim\textsuperscript{1},
  Sangjin Park\textsuperscript{1}\textsuperscript{*} \\
  \textsuperscript{1}Graduate School of Industrial Data Engineering, Hanyang University, Seoul, Republic of Korea \\
  {\tt dnddl9456@hanyang.ac.kr, psj3493@hanyang.ac.kr} \\
  \small{\textsuperscript{*}Corresponding author}
}

\begin{document}
\maketitle
\begin{abstract}
Self-attention encoders such as Bidirectional Encoder Representations from Transformers (BERT) scale quadratically with sequence length, making long-context modeling expensive. Linear-time state-space models, such as Mamba, are efficient; however, they show limitations in modeling global interactions and can suffer from padding-induced state contamination. We propose MaBERT, a hybrid encoder that interleaves Transformer layers for global dependency modeling with Mamba layers for linear-time state updates. This design alternates global contextual integration with fast state accumulation, enabling efficient training and inference on long inputs. To stabilize variable-length batching, we introduce padding-safe masking, which blocks state propagation through padded positions, and mask-aware attention pooling, which aggregates information only from valid tokens. On GLUE, MaBERT achieves the best mean score on five of the eight tasks, with strong performance on the CoLA and sentence-pair inference tasks. When extending the context from 512 to 4,096 tokens, MaBERT reduces training time and inference latency by 2.36$\times$ and 2.43$\times$, respectively, relative to the average of encoder baselines, demonstrating a practical long-context-efficient encoder.
\end{abstract}

\section{Introduction}
Pretrained encoders are central to modern natural language processing and related sequence-modeling tasks, where downstream performance often depends on the quality of input-sequence representations~\cite{RN1}. In Transformer encoder-decoder architectures, the decoder repeatedly queries encoder outputs via cross-attention; therefore, limitations in encoder representations can bottleneck end-to-end quality, even when decoder capacity increases~\cite{RN2,RN3}. Accordingly, pretrained encoders led by  Bidirectional Encoder Representations from Transformers (BERT) have become standard backbones across diverse natural language understanding and time-series applications, offering practical advantages in training and deployment efficiency compared with recent large language models~\cite{RN4,RN5,RN6,RN7,RN8}.

Despite their broad utility, self-attention introduces a critical efficiency bottleneck. Computing all token-pair interactions yields $O(n^2)$ complexity with respect to sequence length, which severely constrains long-context scalability~\cite{RN2,RN9,RN10}. Prior efforts have improved pretraining  strategies or attention designs (e.g., RoBERTa and DeBERTa)~\cite{RN11,RN12} and introduced sparse attention mechanisms to extend context (e.g., Longformer and BigBird)~\cite{RN10,RN13}. However, these methods either restrict global context capture or remain within self-attention-based variants, leaving the fundamental length-dependent cost growth unresolved~\cite{RN14}.

State-Space Models (SSMs) provide a promising alternative by modeling long-range dependencies with linear complexity $O(n)$, compressing sequences into fixed-size hidden states and propagating them over time~\cite{RN15}. Mamba further strengthens SSMs via selective scanning, which adaptively retains or forgets information conditioned on input, thereby improving context-dependent inference~\cite{RN16}. This sequential state-update mechanism is complementary to Transformers' global contextual modeling, motivating hybrid designs that interleave the two at the layer level to jointly achieve efficiency and expressiveness~\cite{RN2,RN17,RN18}.

However, applying such hybrids to bidirectional encoder pretraining with masked language modeling (MLM) reveals a key obstacle. Variable-length batching requires padding, and padding tokens can continue to drive sequential state updates in SSM layers, leading to padding-induced state contamination that distorts valid-token representations ~\cite{RN7,RN16,RN19,RN20,RN21}. Unlike decoders with causal masking, encoders must integrate information from all tokens to form a bidirectional context; therefore, these distortions can propagate through residual paths and degrade sentence-level representations~\cite{RN2}.

To address this issue while maintaining high accuracy and long-context efficiency, we propose MaBERT, a hybrid encoder that integrates Transformer-based global dependency modeling with Mamba-based linear-time sequential updates within a single stack. MaBERT interleaves Transformer self-attention and Mamba layers, alternating between global contextual interactions and efficient state accumulation. To ensure robustness under variable-length inputs, we introduce padding-safe masking (PSM) to block padding-driven state propagation and adopt mask-aware attention pooling (MAP) to aggregate information only from valid tokens, thereby producing stable sentence representations across input lengths. The main contributions of this study are as follows.

\begin{itemize}
  \item We propose MaBERT, an MLM-pretrained hybrid encoder that interleaves Transformer and Mamba layers to combine bidirectional context modeling with linear-time sequential updates.
  \item We address padding-induced state contamination in SSM layers using PSM and MAP, enabling stable representations under variable-length inputs.
  \item MaBERT outperforms strong BERT-family baselines on GLUE (best on 5/8 tasks) and achieves $2.36\times$ faster training and $2.43\times$ lower inference latency when extending context from 512 to 4{,}096 tokens.
\end{itemize}

The remainder of this work is organized as follows: Section~\ref{sec:related_work} reviews related work, Section~\ref{sec:method} describes the MaBERT architecture, Section~\ref{sec:experiments} presents the experimental setup and results, and Section~\ref{sec:conclusion} concludes the paper.

\section{Related Work}\label{sec:related_work}
\subsection{Transformer Encoder Models}
BERT~\cite{RN7} established MLM-based bidirectional pretraining as a strong foundation for encoder representations. Subsequent work improved either representation quality or efficiency: RoBERTa~\cite{RN11} refined the pretraining recipe without architectural changes, DeBERTa~\cite{RN12} enhanced attention via disentangled content and relative position modeling, and ALBERT~\cite{RN23} reduced parameter and memory costs through factorized embeddings and cross-layer sharing. Despite these advances, self-attention retains quadratic cost $O(n^2)$ with respect to sequence length~\cite{RN2}. Long-context variants such as Longformer and BigBird~\cite{RN10,RN13} reduce cost via sparse patterns; however, they restrict interaction structure and do not fully eliminate length-dependent growth in computation and memory~\cite{RN14}. Recent encoders improve long-context efficiency via architectural refinements and system-level optimizations such as kernel accelerations and packing~\cite{dao2022flashattention,krell2021packing,warner2025modernbert}; hybrid encoders instead emphasize attention--SSM interleaving and padding-robust state handling.

\subsection{SSMs}
SSMs provide linear-time $O(n)$ sequence processing by maintaining and updating hidden states~\cite{RN24}. Early linear time-invariant SSMs used input-independent transitions, which limited context-dependent selection~\cite{RN25}. S4~\cite{RN15} enabled stable and efficient training through structured parameterization, and H3~\cite{RN26} introduced gating mechanisms around SSM operations to improve expressiveness. Mamba~\cite{RN16} further advanced SSMs with selective scanning, making state updates input-dependent and achieving strong long-range modeling with hardware-efficient linear-time inference. However, most validations have focused on causal decoders, leaving open questions regarding encoder-style MLM pretraining~\cite{RN24,RN27}.

\begin{figure*}[h]
  \centering
  \includegraphics[width=0.82\linewidth]{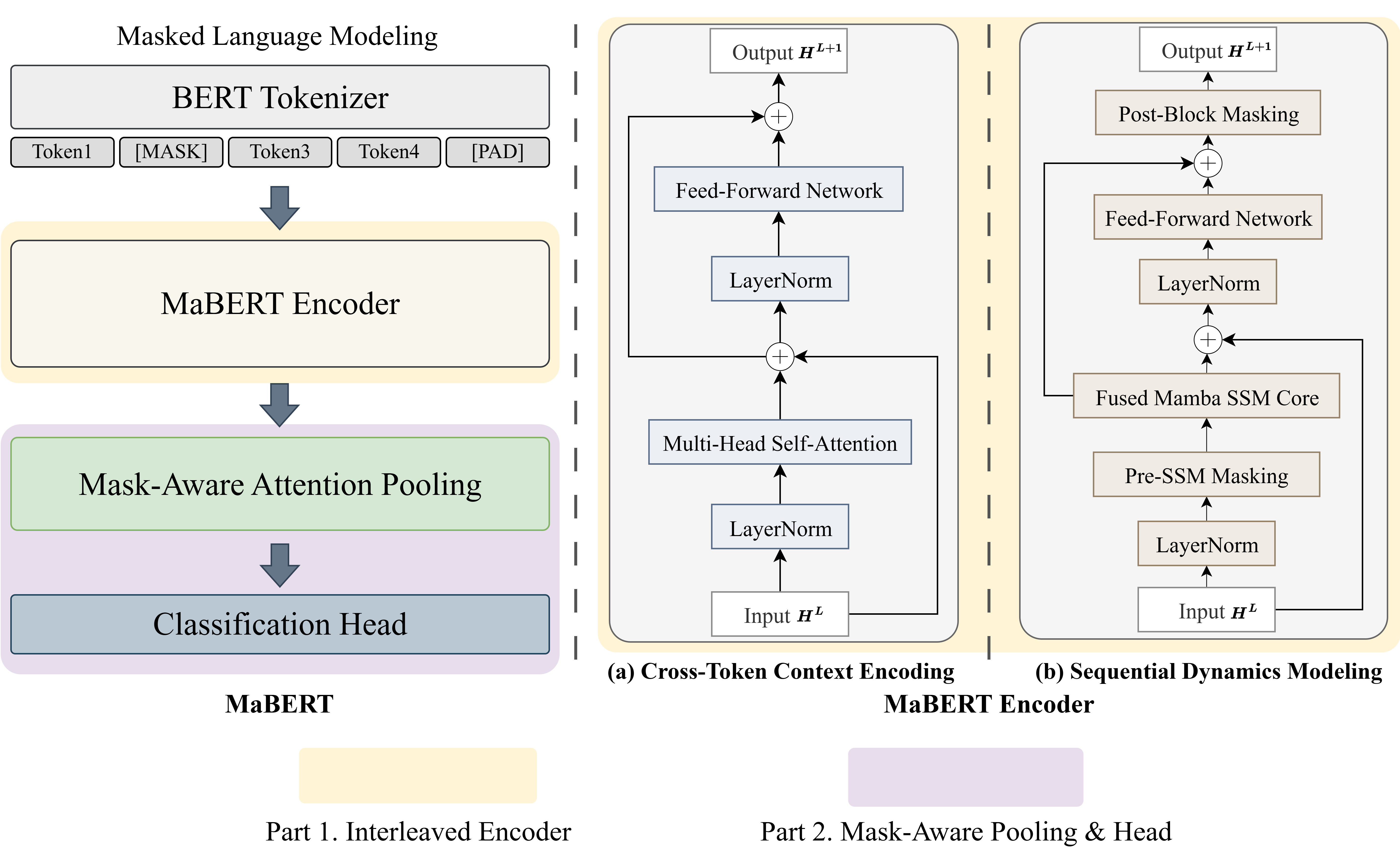}
  \caption{Overall architecture of MaBERT.}
  \label{fig:mabert_overall}
\end{figure*}

\subsection{Hybrid Attention--SSM Models}
Recent hybrids have combined attention-based global interactions with SSM efficiency~\cite{RN24,RN28}. Jamba~\cite{RN17} interleaves the Transformer and Mamba blocks to scale context length, while Hymba~\cite{RN29} couples attention and SSM computations within a layer via hybrid heads, and Nemotron-H~\cite{RN30} explores scaling and mixing strategies with hardware-friendly kernels. These models primarily target causal generation with masking~\cite{RN17,RN29}. In encoder-MLM settings, variable-length batching introduces padding tokens, and sequential SSM updates can accumulate over padding, causing state contamination that degrades valid-token representations~\cite{RN19,RN20}. This has motivated encoder-oriented hybrids that explicitly prevent padding-driven noise in state updates and representation aggregation~\cite{RN7,RN27}.

\section{MaBERT}\label{sec:method}

This section presents MaBERT, an encoder-only hybrid backbone that interleaves Transformer and Mamba layers to combine global self-attention with linear-time state-space updates for long-sequence modeling, as described in Figure~\ref{fig:mabert_overall}. We also describe PSM for variable-length batching and MAP for sentence representation.

\noindent\textbf{(Part 1) Interleaved Encoder:}
MaBERT alternates between global interaction modeling and sequential state accumulation by interleaving Transformer and Mamba layers; Figures~\ref{fig:mabert_overall}(a) and (b) illustrate the computations of the two blocks.

\noindent\textbf{(Part 2) MAP and Head:}
MAP incorporates the padding mask to aggregate sentence representations from valid tokens only, and the resulting vector is fed into a classification head for downstream prediction.

\subsection{Interleaved Encoder}
Part~1 of Figure~\ref{fig:mabert_overall} shows MaBERT, a 12-layer~\cite{RN7} encoder that interleaves Transformer and Mamba blocks to combine global token interactions with sequential processing. We adopt an MMT (Mamba--Mamba--Transformer) schedule repeated four times, which provides the best performance--efficiency trade-off in Section~\ref{sec:pattern_analysis}.

To stabilize heterogeneous interleaving, MaBERT uses a unified Pre-LN residual update scheme~\cite{RN31}. Each block applies LN to its input, performs its sub-operations on the normalized representations, and adds the result back via a residual connection, helping maintain stable training across block types (Figure~\ref{fig:mabert_overall}(a,b)).

\subsection{Cross-Token Context Encoding}
Figure~\ref{fig:mabert_overall}(a) illustrates the Transformer layer for cross-token context encoding. This module models global token-to-token interactions via self-attention and updates each token representation to reflect sentence-level context. Within MaBERT's interleaved design, the Transformer layers periodically re-inject global contextual consistency, whereas sequential updates accumulate in the Mamba layers. 

Let $L$ denote the layer index and $H^{L}\in\mathbb{R}^{B\times T\times D}$ be the input token representations to the $L$-th layer, where $B$, $T$ and $D$ indicate the batch size, the sequence length, and the hidden size, respectively. MaBERT applies a Pre-LN residual update for the Transformer layer, as follows:
\begin{equation}
\begin{gathered}
\bar{H}^{L} = \mathrm{LN}(H^{L}), \\
H_{\text{att}}^{L} = H^{L} + \mathrm{MHSA}(\bar{H}^{L}), \\
H^{L+1} = H_{\text{att}}^{L}
+ \mathrm{FFN}\!\bigl(\mathrm{LN}(H_{\text{att}}^{L})\bigr).
\end{gathered}
\label{eq:transformer_preln}
\end{equation}
Here, $\mathrm{LN}(\cdot)$ denotes Layer Normalization, $\mathrm{MHSA}(\cdot)$ denotes multi-head self-attention, and $\mathrm{FFN}(\cdot)$ denotes a position-wise feed-forward network. Self-attention constructs queries, keys, and values via learned linear projections and applies an additive padding mask to the attention logits so that pad positions do not contribute to softmax normalization. The FFN then enhances token-wise expressiveness through a nonlinear transformation. This cross-token context encoding injects global contextual information into each token representation, and the subsequent Mamba layer further updates these representations by accumulating sequential information with linear-time complexity.

\subsection{Sequential Dynamics Modeling}
\label{sec:sequential_dynamics}
Figure~\ref{fig:mabert_overall}(b) illustrates the sequential dynamics modeling of the fused Mamba SSM core in a MaBERT encoder layer, which updates token representations in linear time with respect to the sequence length $T$. Under variable-length batching during encoder pretraining, padding tokens can still drive sequential state updates and contaminate the internal state, thereby distorting valid-token representations. To prevent this, MaBERT applies PSM both immediately before the fused SSM core (Pre-SSM Masking) and at the block output (Post-Block Masking), as shown in Figure~\ref{fig:mabert_overall}(b).

We define the layer input as $H^{L}\in\mathbb{R}^{B\times T\times D}$ and the token representation at position $t$ as $h_t^{L}\in\mathbb{R}^{B\times D}$. Here, $m\in\{0,1\}^{B\times T}$ denotes the padding mask, with $\tilde{m}\in\{0,1\}^{B\times T\times 1}$ as its last-dimension expansion broadcast over the hidden dimension in element-wise products. Following Figure~\ref{fig:mabert_overall}(b), MaBERT adopts a Pre-LN residual structure: after LayerNorm, the masked input is fed to the SSM core and accumulated via a residual addition.
\begin{equation}
\begin{gathered}
\bar{H}^{L} = \mathrm{LN}(H^{L}), \\
\hat{H}^{L} = \tilde{m}\odot \bar{H}^{L}, \\
H_{\mathrm{ssm}}^{L} = H^{L} + \mathrm{SSM}(\hat{H}^{L}).
\end{gathered}
\label{eq:mamba_pre_mask}
\end{equation}
The layer then applies the FFN and re-applies masking at the output:
\begin{equation}
\small
\begin{aligned}
H^{L+1}
&= \tilde{m}\odot\Bigl(
H_{\mathrm{ssm}}^{L}
+ \mathrm{FFN}\!\bigl(\mathrm{LN}(H_{\mathrm{ssm}}^{L})\bigr)
\Bigr).
\end{aligned}
\label{eq:mamba_post_mask}
\end{equation}
This two-stage design is necessary because residual paths and the FFN can reintroduce nonzero values at padded positions even if the SSM input is masked; post-block masking re-zeros them so they do not persist as inputs to upper layers (e.g., LN/residual), reducing length-dependent drift.

\paragraph{(i) Input--Gate Split and Local Mixing.}
The SSM core takes the normalized and masked token representation $\hat{h}_t^{L}$ and splits it into an input path and a gating path. Internally, the computation dimension is expanded to $D_m=\varepsilon D$, where $\varepsilon$ is the expansion ratio. Using a learnable projection $W_{\mathrm{in}}\in\mathbb{R}^{D\times 2D_m}$, we form
\begin{equation}
\begin{gathered}
[u_t,z_t] = \hat{h}_t^{L}W_{\mathrm{in}}, \\
U = [u_1;\ldots;u_T]\in\mathbb{R}^{B\times T\times D_m},
\end{gathered}
\label{eq:split}
\end{equation}
where $u_t,z_t\in\mathbb{R}^{B\times D_m}$ denote the input and gating paths, respectively. To incorporate local context, we apply a depth-wise one-dimensional convolution along the sequence dimension to $U$:
\begin{equation}
\begin{gathered}
\tilde{U} = \mathrm{DWConv}(U), \\
\tilde{u}_t = \tilde{U}[:,t,:].
\end{gathered}
\label{eq:dwconv}
\end{equation}
Here, $\mathrm{DWConv}(\cdot)$ processes each channel independently and is implemented along the sequence dimension following the Mamba block design to support efficient sequential computation.

\paragraph{(ii) Token-wise Parameterization.}
In selective SSMs, position-specific coefficients are generated and conditioned on the input. Given $\tilde{u}_t$, MaBERT produces a low-rank representation $d_t$ for step-size generation, an input-injection coefficient $b_t$, and an output coefficient $c_t$. Using $W_x\in\mathbb{R}^{D_m\times (r+2N)}$, where $r$ is the $\Delta$-rank and $N$ is the state expansion dimension:
\begin{equation}
\begin{gathered}
[d_t,b_t,c_t] = \tilde{u}_t W_x, \\
d_t \in \mathbb{R}^{B\times r}, \qquad
b_t,c_t \in \mathbb{R}^{B\times N}.
\end{gathered}
\label{eq:coeffs}
\end{equation}
The step size $\Delta_t\in\mathbb{R}^{B\times D_m}$ is obtained by re-projecting $d_t$ with $W_{\Delta}\in\mathbb{R}^{r\times D_m}$, adding a bias, and applying $\mathrm{softplus}$ to ensure positivity:
\begin{equation}
\Delta_t=\mathrm{softplus}(d_t W_{\Delta}+b_{\Delta}).
\label{eq:delta}
\end{equation}

\paragraph{(iii) Selective State Update and Gated Readout.}
We use a channel-wise diagonal transition and parameterize the transition matrix $A\in\mathbb{R}^{D_m\times N}$ in the negative domain to ensure stable decaying dynamics:
\begin{equation}
A=-\exp(A_{\log}).
\label{eq:A}
\end{equation}
A learnable channel-wise skip connection $D_{\mathrm{skip}}\in\mathbb{R}^{D_m}$ is also included. The selective scan sequentially updates the internal state using $\Delta_t,A,b_t,c_t$ together with the input-path signal $\tilde{u}_t$, and forms the output by reading out along the state dimension via $c_t$. Given the scan readout $\tilde{o}_t$, the gating path applies a nonlinear gate defined as $o_t=\mathrm{SiLU}(z_t)\odot \tilde{o}_t$, where $\mathrm{SiLU}(x)=x\cdot \sigma(x)$ and $\sigma(\cdot)$ denotes the sigmoid function. Finally, outputs are projected back to the original hidden size using $W_{\mathrm{out}}\in\mathbb{R}^{D_m\times D}$:
\begin{equation}
\begin{gathered}
y_t = o_t W_{\mathrm{out}}, \\
Y = [y_1;\ldots;y_T]\in\mathbb{R}^{B\times T\times D}.
\end{gathered}
\label{eq:outproj}
\end{equation}

\paragraph{(iv) PSM in Variable-Length Batches.}
Even with end-padding, padding can affect \emph{boundary} valid tokens through local mixing (e.g., DWConv) inside Mamba blocks; subsequent Transformer layers may then spread this boundary noise globally. Figure~\ref{fig:mabert_overall}(b) shows a two-stage PSM: (1) pre-SSM masking blocks padding activations from entering sequential updates (Eq.~\ref{eq:mamba_pre_mask}); and (2) post-block masking re-zeros pad outputs after residual/FFN so they do not persist to upper layers (Eq.~\ref{eq:mamba_post_mask}). At the token level,
\begin{equation}
\small
\begin{gathered}
\hat{h}_t^{L} = m_t\odot \bar{h}_t^{L}, \\
h_t^{L+1} = m_t\odot\Bigl(
h_{t,\mathrm{ssm}}^{L}
+ \mathrm{FFN}\!\bigl(\mathrm{LN}(h_{t,\mathrm{ssm}}^{L})\bigr)
\Bigr),
\end{gathered}
\label{eq:token_level_mask}
\end{equation}
where $\bar{h}_t^{L}=\mathrm{LN}(h_t^{L})$ and $h_{t,\mathrm{ssm}}^{L}$ denote the token representation at position $t$ in $H_{\mathrm{ssm}}^{L}$. This padding-safe treatment suppresses padding-driven state contamination and stabilizes representation learning under variable-length inputs.

\begin{table*}[h]
  \centering
  \small
  \setlength{\tabcolsep}{6pt}
  \renewcommand\cellalign{cc}
  \begin{tabular}{lcccccccc}
    \hline
    \textbf{Encoder pattern} &
    \textbf{CoLA} & \textbf{SST-2} & \textbf{MRPC} & \textbf{QQP} &
    \textbf{MNLI-m} & \textbf{MNLI-mm} & \textbf{QNLI} & \textbf{RTE} \\
    \hline
    MMMMMMMMMMMM &
    \makecell{0.401\\$\pm$0.014} &
    \makecell{0.878\\$\pm$0.007} &
    \makecell{0.805\\$\pm$0.017} &
    \makecell{0.831\\$\pm$0.003} &
    \makecell{0.745\\$\pm$0.012} &
    \makecell{0.740\\$\pm$0.017} &
    \makecell{0.796\\$\pm$0.008} &
    \makecell{0.561\\$\pm$0.032} \\
    TTTTTTTTTTTT &
    \makecell{0.428\\$\pm$0.020} &
    \makecell{0.891\\$\pm$0.012} &
    \makecell{0.834\\$\pm$0.015} &
    \makecell{0.843\\$\pm$0.002} &
    \makecell{0.796\\$\pm$0.015} &
    \makecell{0.802\\$\pm$0.018} &
    \makecell{0.864\\$\pm$0.007} &
    \makecell{0.575\\$\pm$0.030} \\
    MTTMTTMTTMTT &
    \makecell{0.555\\$\pm$0.017} &
    \makecell{0.913\\$\pm$0.008} &
    \makecell{0.823\\$\pm$0.015} &
    \makecell{0.855\\$\pm$0.003} &
    \makecell{0.804\\$\pm$0.015} &
    \makecell{0.806\\$\pm$0.019} &
    \makecell{0.868\\$\pm$0.008} &
    \makecell{0.586\\$\pm$0.030} \\
    TMTMTMTMTMTM &
    \makecell{0.525\\$\pm$0.017} &
    \makecell{0.896\\$\pm$0.011} &
    \makecell{0.826\\$\pm$0.017} &
    \makecell{0.862\\$\pm$0.004} &
    \makecell{0.801\\$\pm$0.014} &
    \makecell{0.802\\$\pm$0.018} &
    \makecell{0.873\\$\pm$0.006} &
    \makecell{0.587\\$\pm$0.031} \\
    MTMTMTMTMTMT &
    \makecell{0.573\\$\pm$0.019} &
    \makecell{0.897\\$\pm$0.010} &
    \makecell{0.797\\$\pm$0.018} &
    \makecell{0.861\\$\pm$0.003} &
    \makecell{0.803\\$\pm$0.015} &
    \makecell{0.804\\$\pm$0.016} &
    \makecell{0.859\\$\pm$0.007} &
    \makecell{0.582\\$\pm$0.031} \\
    TMMTMMTMMTMM &
    \makecell{0.528\\$\pm$0.014} &
    \makecell{0.903\\$\pm$0.008} &
    \makecell{0.832\\$\pm$0.016} &
    \makecell{0.863\\$\pm$0.004} &
    \makecell{0.803\\$\pm$0.014} &
    \makecell{0.802\\$\pm$0.026} &
    \makecell{0.870\\$\pm$0.009} &
    \makecell{0.595\\$\pm$0.031} \\
    \textbf{MMTMMTMMTMMT} &
    \makecell{\textbf{0.574}\\$\pm$0.016} &
    \makecell{\textbf{0.904}\\$\pm$0.009} &
    \makecell{\textbf{0.837}\\$\pm$0.016} &
    \makecell{\textbf{0.868}\\$\pm$0.003} &
    \makecell{\textbf{0.809}\\$\pm$0.014} &
    \makecell{\textbf{0.814}\\$\pm$0.015} &
    \makecell{0.867\\$\pm$0.007} &
    \makecell{\textbf{0.602}\\$\pm$0.030} \\
    TTMTTMTTMTTM &
    \makecell{0.512\\$\pm$0.016} &
    \makecell{0.902\\$\pm$0.009} &
    \makecell{0.809\\$\pm$0.016} &
    \makecell{0.859\\$\pm$0.003} &
    \makecell{0.800\\$\pm$0.014} &
    \makecell{0.805\\$\pm$0.020} &
    \makecell{\textbf{0.874}\\$\pm$0.007} &
    \makecell{0.580\\$\pm$0.030} \\
    \hline
  \end{tabular}
  \caption{GLUE benchmark scores across interleaved Transformer--Mamba encoder patterns. Models are pretrained with 10\% of total steps.
  M and T denote Mamba and Transformer layers, respectively.
  CoLA uses Matthews correlation coefficient; SST-2, MNLI-m, MNLI-mm, QNLI, and RTE use accuracy; MRPC and QQP use F1.}
  \label{tab:pattern_glue}
\end{table*}

\subsection{MAP and Head}
In Part~2 of Figure~\ref{fig:mabert_overall}, MaBERT 
forms a sentence-level representation from the encoder's final token embeddings and maps it to downstream predictions. Rather than relying solely on a single [CLS] token, MaBERT uses MAP, which explicitly excludes padded tokens while assigning higher weights to semantically informative tokens. This design prevents the padded regions from distorting sentence representations and yields robust aggregation when information is distributed across multiple positions.

Let the encoder's final output be $H\in\mathbb{R}^{B\times T\times D}$, where $H_t\in\mathbb{R}^{B\times D}$ denotes the token embedding at position $t$. We compute the token scores via a linear projection:
\begin{equation}
p_t = H_t W_s,\qquad W_s\in\mathbb{R}^{D\times 1},
\label{eq:pool_score}
\end{equation}
and stack them to obtain $p\in\mathbb{R}^{B\times T\times 1}$. To ensure that the padded tokens receive zero weight, we apply a masked softmax by adding a large negative constant to the masked locations:
\begin{equation}
\alpha
= \mathrm{softmax}\!\left(
p + (1-\tilde{m})\cdot(-\kappa)
\right),
\label{eq:masked_softmax}
\end{equation}
where $\kappa$ is a sufficiently large positive constant, and $\alpha\in\mathbb{R}^{B\times T\times 1}$ denotes normalized attention weights.

\begin{figure}[t]
  \centering
  \includegraphics[width=\linewidth]{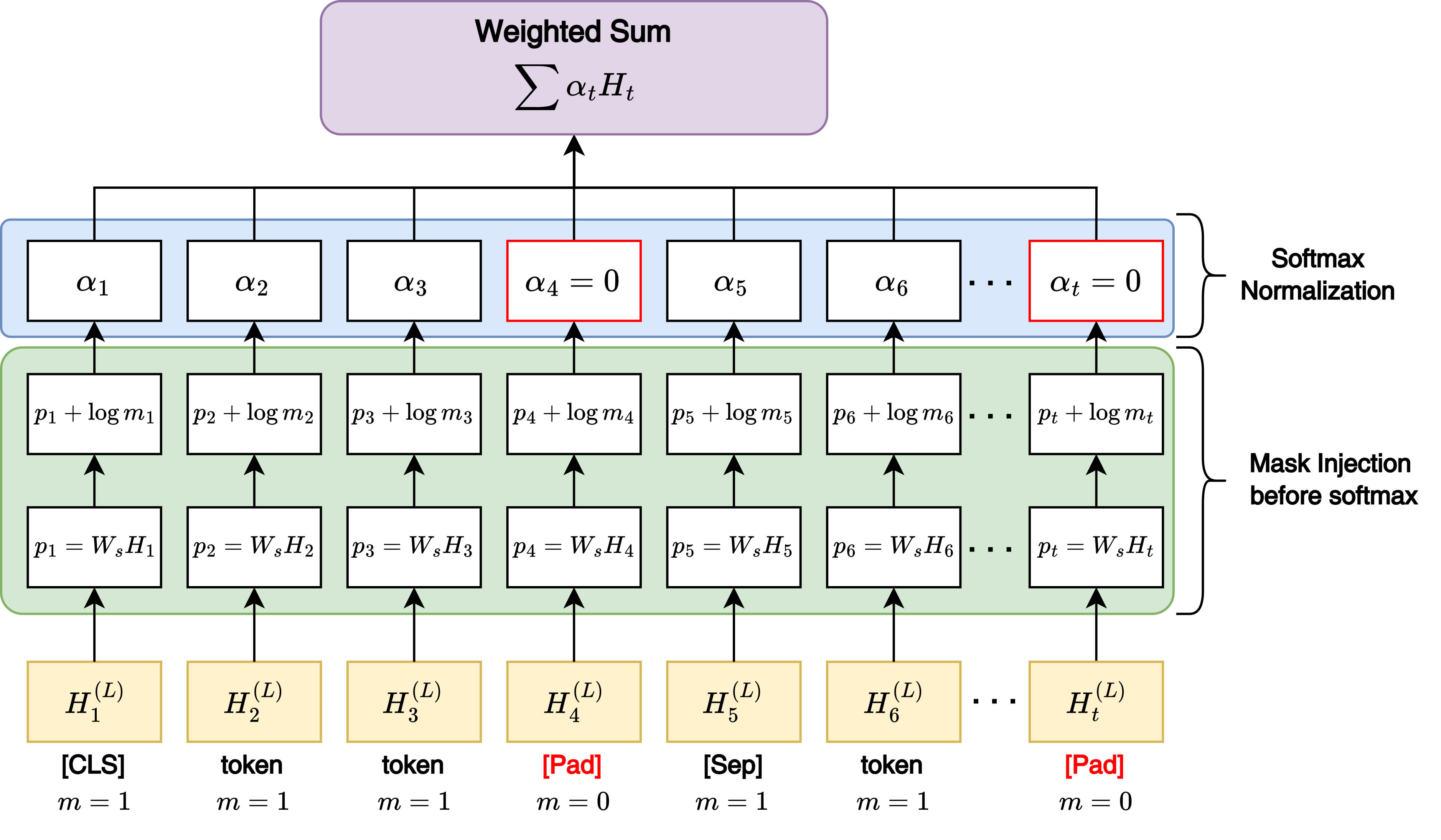}
  \caption{Mask-aware attention pooling in MaBERT.}
  \label{fig:mapooling}
\end{figure}

As illustrated in Figure~\ref{fig:mapooling}, MAP computes the token scores, injects the padding mask before normalization, and aggregates only valid-token representations through a weighted sum. The pooled sentence representation is computed as follows:
\begin{equation}
h_{\mathrm{pool}}
= \sum_{t=1}^{T}\alpha_t\,H_t,
\qquad
h_{\mathrm{pool}}\in\mathbb{R}^{B\times D}.
\label{eq:pool_sum}
\end{equation}
It is then passed to a classification head:
\begin{equation}
\hat{y}=W_c\,\mathrm{Dropout}(h_{\mathrm{pool}})+b_c,
\label{eq:cls_head}
\end{equation}
where $W_c\in\mathbb{R}^{C\times D}$ and $b_c\in\mathbb{R}^{C}$ are learnable parameters, $\hat{y}$ denotes class logits, and $C$ is the number of classes.

\section{Experiments}\label{sec:experiments}
\label{sec:experiments}

This section presents the experimental setup and GLUE results, followed by in-depth analyses of interleaving patterns, pretraining budgets, ablations, and efficiency and scalability.

\subsection{Experimental Setup}
\label{sec:exp_setup}

\paragraph{Baselines.}
We compare MaBERT against BERT~\cite{RN7}, ALBERT~\cite{RN23}, BigBird~\cite{RN13}, Longformer~\cite{RN10}, and DeBERTa~\cite{RN12} under a matched MLM pretraining recipe to isolate architecture-dependent differences. Each model uses its default tokenizer; all other pretraining settings and compute budgets are aligned (Table~\ref{tab:appendix_recipe}).

\paragraph{Dataset.}
We evaluate eight GLUE tasks (CoLA, SST-2, MRPC, QQP, MNLI-m/mm, QNLI, RTE)~\cite{RN32} with official metrics: MCC for CoLA, accuracy for SST-2/MNLI-m/MNLI-mm/QNLI/RTE, and accuracy and F1 for MRPC and QQP.

\paragraph{Protocol.}
All models are pretrained on BookCorpus~\cite{RN33} and English Wikipedia using MLM only. We match the BERT-based 1M-step budget~\cite{RN7} across models and report results at 10\%, 25\%, 50\%, and 100\% of steps using a two-stage length schedule (128 then 512 tokens). We report mean and standard deviation over five seeds; the shared configuration is summarized in Table~\ref{tab:appendix_recipe} (Appendix).

\paragraph{Efficiency.}
We measure training-step time, inference latency, and peak memory on a fixed GPU with matched length, batch size, and precision. BERT, ALBERT, and DeBERTa use PyTorch SDPA on CUDA~\cite{RN34}; BigBird and Longformer use their sparse-attention implementations with warm-up and synchronized repeats. Although MaBERT has more parameters at the same 12-layer depth, this increase is inherent to adding SSM capacity; we therefore report how memory and runtime scale with length rather than strict parameter matching, with details (incl.\ 4,096 positional extension) in Table~\ref{tab:appendix_eff_setup} (Appendix). We keep the backend uniform and avoid implementation-dependent optimizations (e.g., FlashAttention) for fair comparisons.

\vspace{-0.8em}

\subsection{Interleaving Pattern Analysis}
\label{sec:pattern_analysis}

\vspace{-0.4em}

We evaluated GLUE performance across Transformer--Mamba interleaving schedules. To screen many candidates efficiently, we used 10\% of the pretraining budget, fixed the encoder depth to 12 layers, and tested eight representative schedules that varied the mixing ratio and placement, including Transformer-only and Mamba-only. Table~\ref{tab:pattern_glue} reports the mean and standard deviation over five seeds for eight GLUE tasks.

\begin{table*}[!t]
  \centering
  \small
  \setlength{\tabcolsep}{6pt}
  \renewcommand\cellalign{cc}
  \begin{tabular}{lcccccccc}
    \hline
    \textbf{Model} &
    \textbf{CoLA} & \textbf{SST-2} & \textbf{MRPC} & \textbf{QQP} &
    \textbf{MNLI-m} & \textbf{MNLI-mm} & \textbf{QNLI} & \textbf{RTE} \\
    \hline
    BERT &
    \makecell{0.522\\$\pm$0.017} &
    \makecell{0.912\\$\pm$0.012} &
    \makecell{0.853\\$\pm$0.017} &
    \makecell{0.856\\$\pm$0.006} &
    \makecell{0.826\\$\pm$0.014} &
    \makecell{0.829\\$\pm$0.011} &
    \makecell{0.876\\$\pm$0.012} &
    \makecell{0.618\\$\pm$0.027} \\
    ALBERT &
    \makecell{0.503\\$\pm$0.018} &
    \makecell{0.920\\$\pm$0.012} &
    \makecell{0.855\\$\pm$0.017} &
    \makecell{0.857\\$\pm$0.006} &
    \makecell{0.829\\$\pm$0.012} &
    \makecell{0.832\\$\pm$0.013} &
    \makecell{0.880\\$\pm$0.013} &
    \makecell{0.618\\$\pm$0.030} \\
    Longformer &
    \makecell{0.534\\$\pm$0.018} &
    \makecell{0.924\\$\pm$0.012} &
    \makecell{0.863\\$\pm$0.016} &
    \makecell{0.858\\$\pm$0.006} &
    \makecell{0.830\\$\pm$0.013} &
    \makecell{0.831\\$\pm$0.014} &
    \makecell{0.882\\$\pm$0.012} &
    \makecell{0.626\\$\pm$0.031} \\
    BigBird &
    \makecell{0.528\\$\pm$0.016} &
    \makecell{0.926\\$\pm$0.011} &
    \makecell{0.864\\$\pm$0.015} &
    \makecell{0.857\\$\pm$0.005} &
    \makecell{0.831\\$\pm$0.014} &
    \makecell{0.832\\$\pm$0.012} &
    \makecell{0.881\\$\pm$0.013} &
    \makecell{0.624\\$\pm$0.029} \\
    DeBERTa &
    \makecell{0.617\\$\pm$0.015} &
    \makecell{\textbf{0.934}\\$\pm$0.013} &
    \makecell{0.862\\$\pm$0.014} &
    \makecell{0.868\\$\pm$0.004} &
    \makecell{\textbf{0.838}\\$\pm$0.015} &
    \makecell{\textbf{0.842}\\$\pm$0.013} &
    \makecell{0.886\\$\pm$0.019} &
    \makecell{0.648\\$\pm$0.034} \\
    \textbf{MaBERT} &
    \makecell{\textbf{0.676}\\$\pm$0.018} &
    \makecell{0.933\\$\pm$0.010} &
    \makecell{\textbf{0.869}\\$\pm$0.017} &
    \makecell{\textbf{0.879}\\$\pm$0.005} &
    \makecell{0.835\\$\pm$0.016} &
    \makecell{0.837\\$\pm$0.017} &
    \makecell{\textbf{0.893}\\$\pm$0.012} &
    \makecell{\textbf{0.654}\\$\pm$0.033} \\
    \hline
  \end{tabular}
  \caption{Performance comparison of baselines and the proposed MaBERT on the GLUE benchmark.
  Results are reported after full-budget pretraining (100\% steps).
  The best result for each task is highlighted in bold.}
  \label{tab:glue_full_budget}
\end{table*}

As shown in Table~\ref{tab:pattern_glue}, single-family patterns consistently underperform mixed schedules: the Mamba-only encoder is worst overall, and Transformer-only improves but is still weaker than interleaved designs on most tasks. Among mixed patterns, \texttt{MMTMMTMMTMMT} performs best overall, ranking highest on CoLA, MRPC, MNLI-m, MNLI-mm, and RTE while remaining competitive on the remaining tasks. We therefore adopt \texttt{MMTMMTMMTMMT} as the default encoder pattern in subsequent experiments.

\subsection{Pretraining Budgets Analysis}
\label{sec:budget_results}

\begin{figure}[h]
  \centering
  \includegraphics[width=0.8\linewidth]{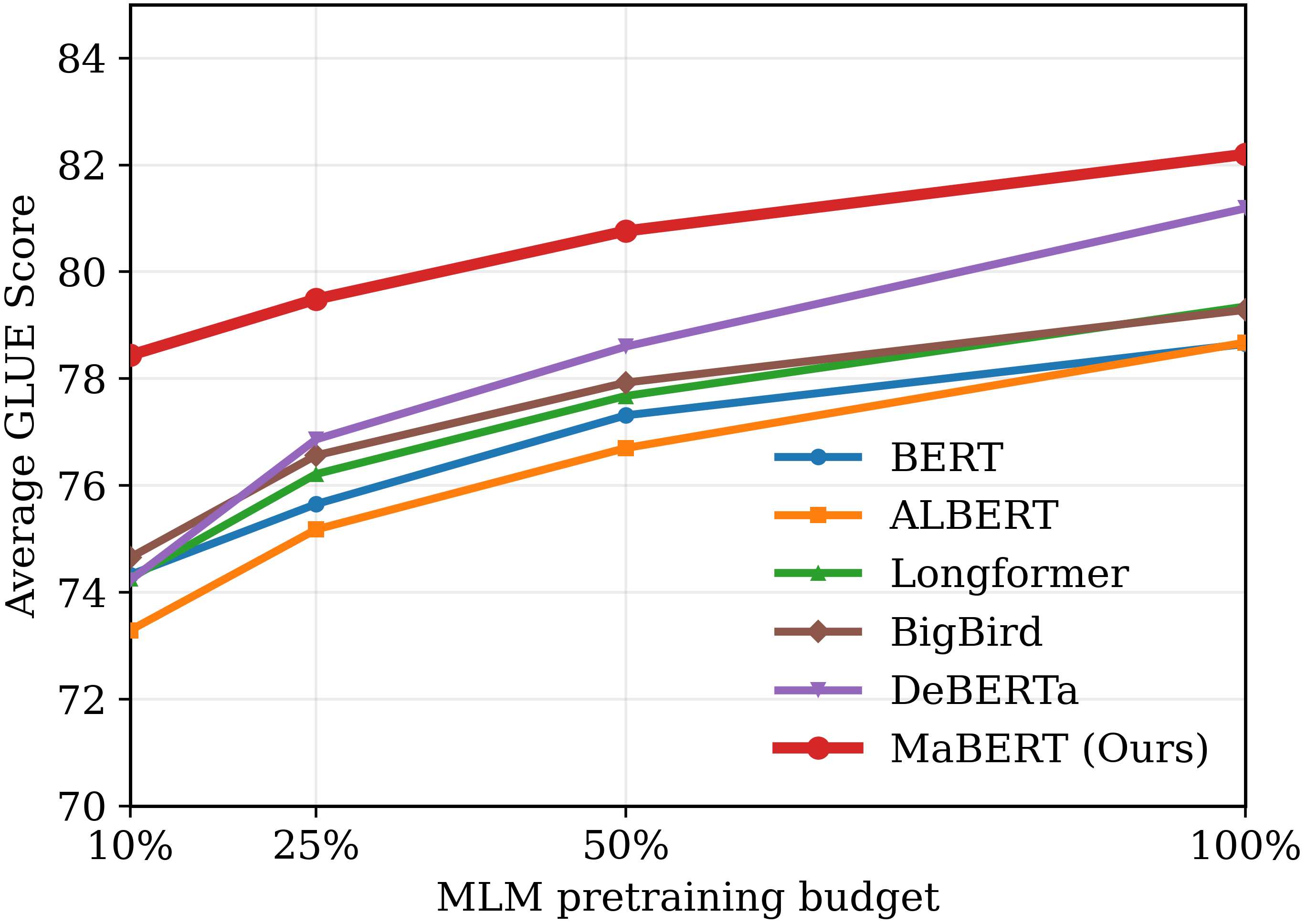}
  \caption{Average GLUE score across pretraining budgets.}
  \label{fig:glue_budget_avg}
\end{figure}

This section compares and analyzes MaBERT using the GLUE benchmark. Figure~\ref{fig:glue_budget_avg} shows the results under pre-training budgets of 10\%, 25\%, 50\%, and 100\%, while all other settings followed the protocol in Section~\ref{sec:exp_setup}.

As shown in Figure~\ref{fig:glue_budget_avg}, MaBERT ranked among the top models in terms of the average GLUE score across all budget regimes and improves steadily as the pretraining budget increases.
Notably, it achieves strong initial performance even in low-budget settings, reaching competitive accuracy with limited pretraining.

Table~\ref{tab:glue_full_budget} presents task-level GLUE results after full-budget pretraining. MaBERT achieves the best performance on CoLA and on several sentence-pair tasks, namely MRPC, QQP, QNLI, and RTE, while remaining competitive on the other tasks. These results suggest that interleaving Transformer layers for global interaction modeling with Mamba layers for sequential state updates enables MaBERT to effectively incorporate sentence-level consistency signals. Full task-wise results across budgets are provided in Tables~\ref{tab:glue_10pct}--\ref{tab:glue_50pct} in the Appendix.

\subsection{Component and Integration Ablations}
\label{sec:ablation}

This section validates the contribution of each component to MaBERT via ablation studies. We first quantify the component-wise effects by comparing the full model and its variants in Table~\ref{tab:ablation_integration}, and then further diagnose the impact of PSM from the perspective of representation stability in Figures~\ref{fig:cola_cosdist_padlen} and~\ref{fig:cola_cosdist_decomp}.

\setcounter{figure}{2}
\begin{figure}[h]
  \centering

  % ---------------- Figure 4 ----------------
  \begin{minipage}{0.9\linewidth}
    \centering
    \begin{subfigure}[b]{0.48\linewidth}
      \centering
      \includegraphics[width=\linewidth]{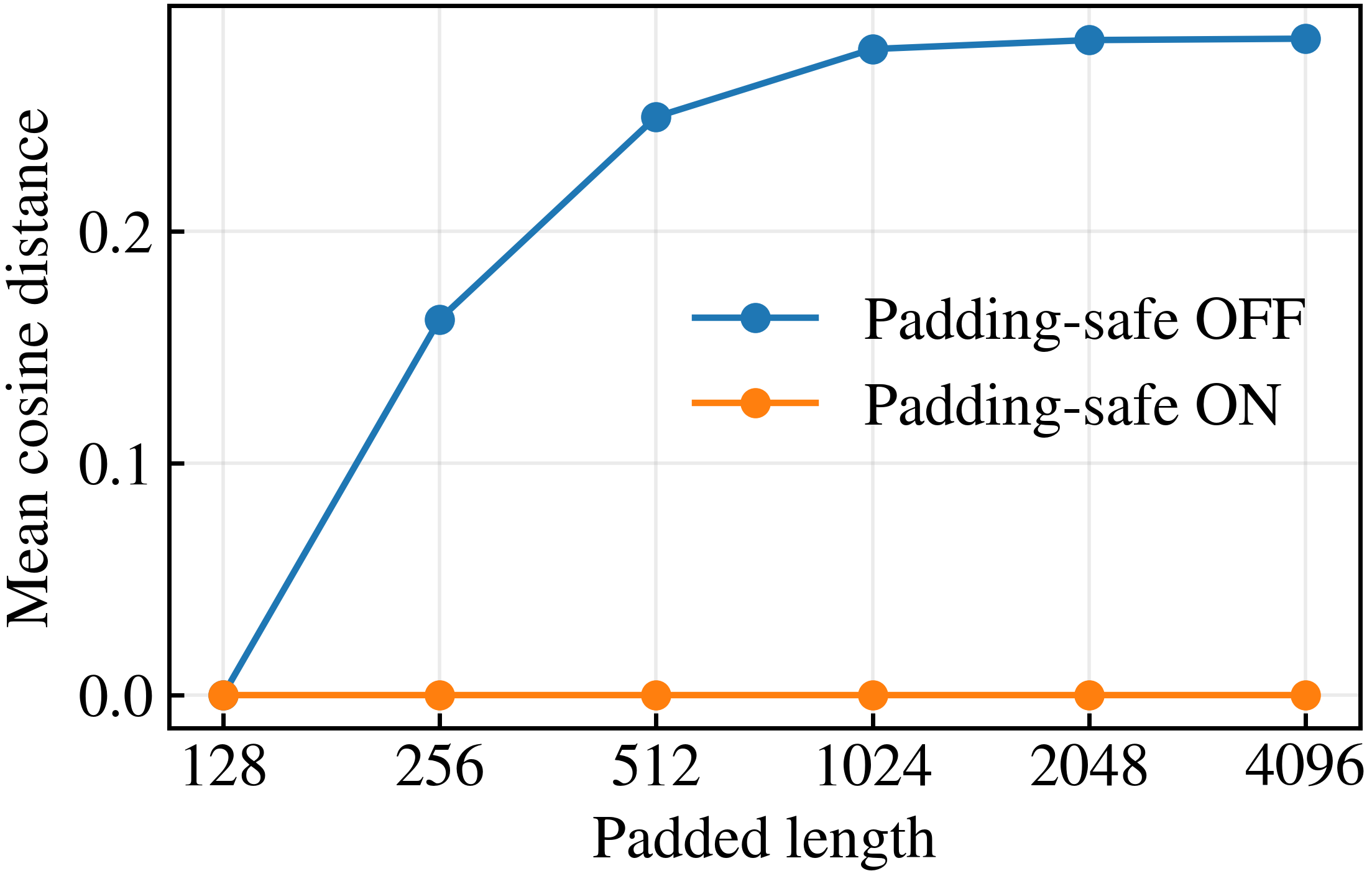}
      \caption{Final}
    \end{subfigure}
    \hfill
    \begin{subfigure}[b]{0.48\linewidth}
      \centering
      \includegraphics[width=\linewidth]{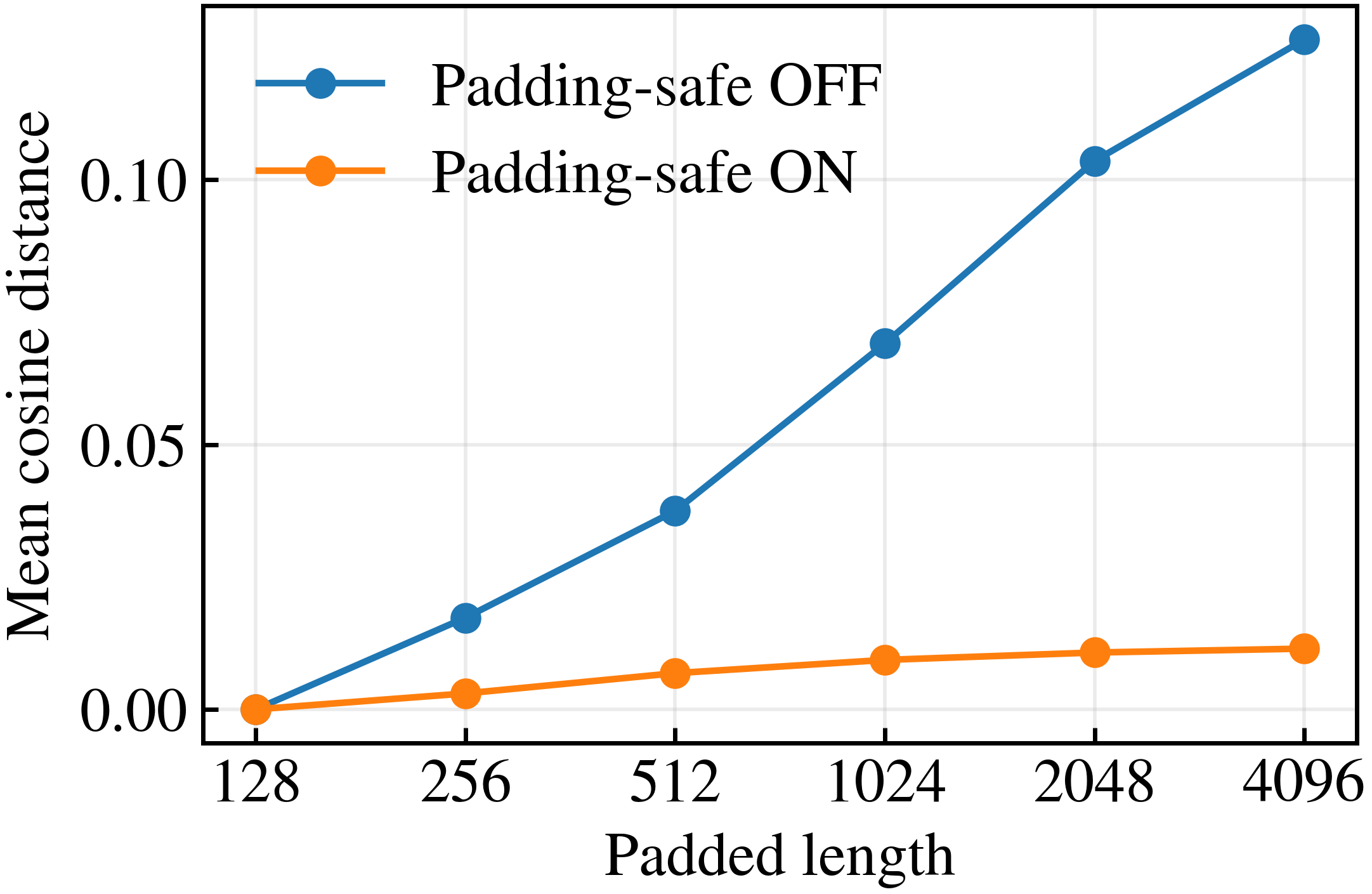}
      \caption{Unmasked mean}
    \end{subfigure}

    \captionof{figure}{Mean cosine distance under padding-length increase on the CoLA dev set.}
    \label{fig:cola_cosdist_padlen}
  \end{minipage}

  \vspace{6pt}

  % ---------------- Figure 5 ----------------
  \begin{minipage}{0.9\linewidth}
    \centering
    \begin{subfigure}[b]{0.48\linewidth}
      \centering
      \includegraphics[width=\linewidth]{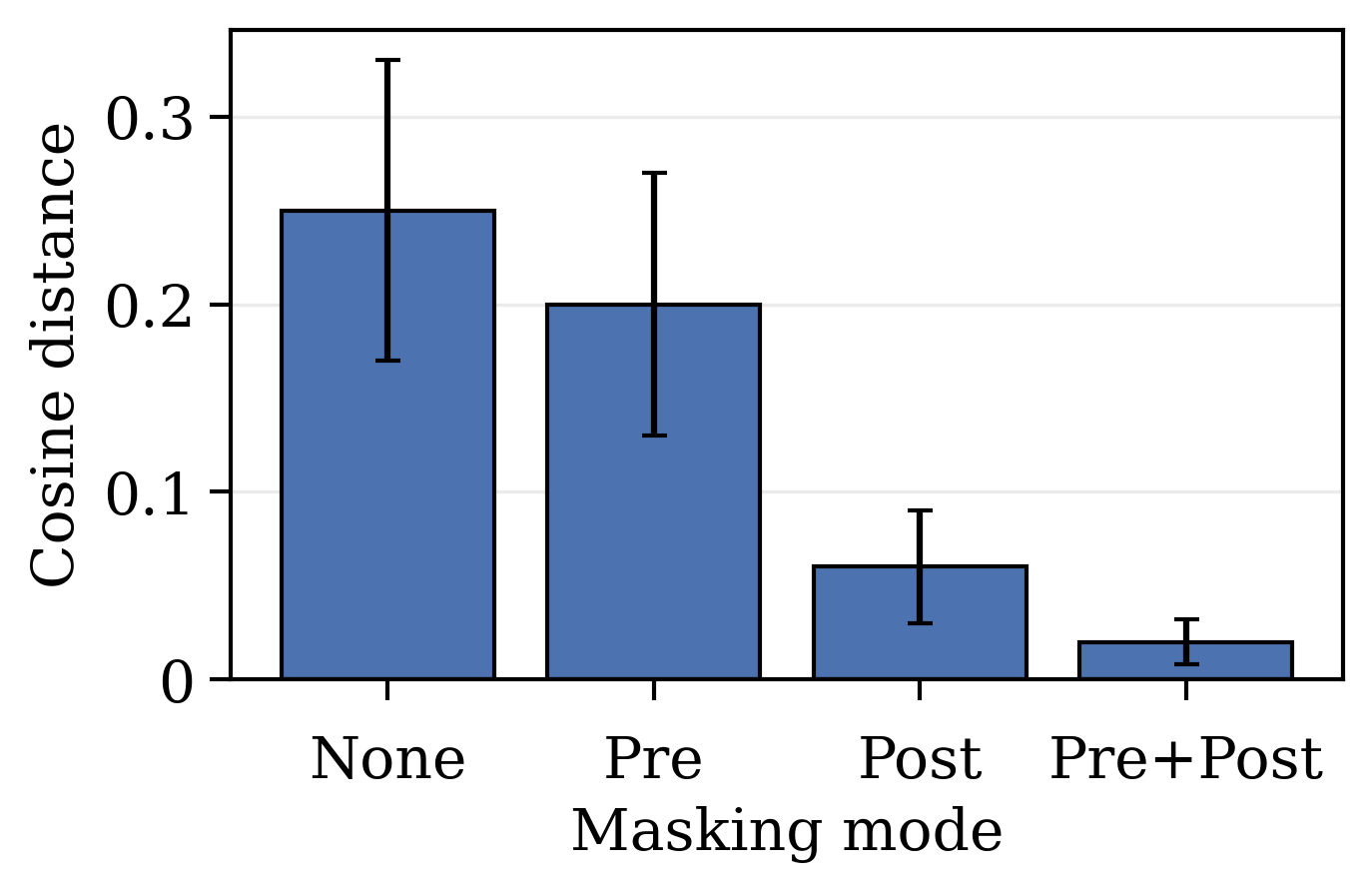}
      \caption{Final}
    \end{subfigure}
    \hfill
    \begin{subfigure}[b]{0.48\linewidth}
      \centering
      \includegraphics[width=\linewidth]{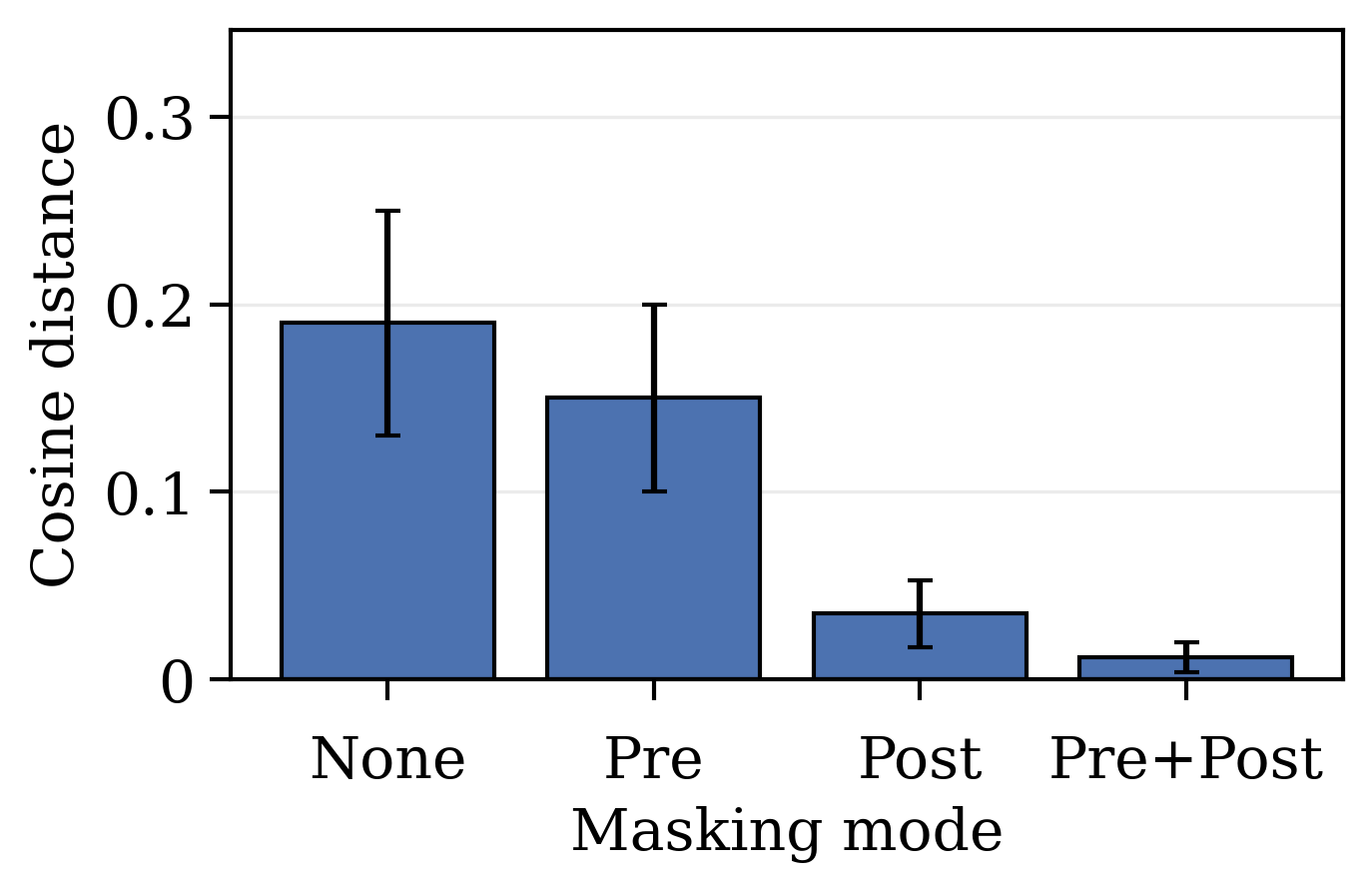}
      \caption{Unmasked mean}
    \end{subfigure}

    \captionof{figure}{Mean cosine distance under padding-length increase on the CoLA dev set, decomposed by padding-safe masking placement.}
    \label{fig:cola_cosdist_decomp}
  \end{minipage}

\end{figure}

\begin{table*}[h]
  \centering
  \small
  \setlength{\tabcolsep}{6pt}
  \renewcommand\cellalign{cc}
  \begin{tabular}{lcccccccc}
    \hline
    \textbf{Model} &
    \textbf{CoLA} & \textbf{SST-2} & \textbf{MRPC} & \textbf{QQP} &
    \textbf{MNLI-m} & \textbf{MNLI-mm} & \textbf{QNLI} & \textbf{RTE} \\
    \hline
    Full &
    \makecell{\textbf{0.676}\\$\pm$0.018} &
    \makecell{\textbf{0.933}\\$\pm$0.010} &
    \makecell{\textbf{0.869}\\$\pm$0.017} &
    \makecell{\textbf{0.879}\\$\pm$0.005} &
    \makecell{\textbf{0.835}\\$\pm$0.016} &
    \makecell{\textbf{0.837}\\$\pm$0.017} &
    \makecell{\textbf{0.893}\\$\pm$0.012} &
    \makecell{\textbf{0.654}\\$\pm$0.033} \\
    PSM only &
    \makecell{0.641\\$\pm$0.022} &
    \makecell{0.918\\$\pm$0.011} &
    \makecell{0.849\\$\pm$0.019} &
    \makecell{0.863\\$\pm$0.006} &
    \makecell{0.816\\$\pm$0.017} &
    \makecell{0.818\\$\pm$0.018} &
    \makecell{0.874\\$\pm$0.013} &
    \makecell{0.638\\$\pm$0.036} \\
    MAP only &
    \makecell{0.661\\$\pm$0.018} &
    \makecell{0.922\\$\pm$0.031} &
    \makecell{0.841\\$\pm$0.023} &
    \makecell{0.860\\$\pm$0.004} &
    \makecell{0.819\\$\pm$0.017} &
    \makecell{0.820\\$\pm$0.027} &
    \makecell{0.878\\$\pm$0.014} &
    \makecell{0.597\\$\pm$0.078} \\
    None &
    \makecell{0.596\\$\pm$0.027} &
    \makecell{0.903\\$\pm$0.013} &
    \makecell{0.841\\$\pm$0.021} &
    \makecell{0.847\\$\pm$0.007} &
    \makecell{0.803\\$\pm$0.020} &
    \makecell{0.805\\$\pm$0.021} &
    \makecell{0.855\\$\pm$0.015} &
    \makecell{0.614\\$\pm$0.041} \\
    \hline
  \end{tabular}
  \caption{Integration ablation results of MaBERT on the GLUE benchmark.
  PSM denotes padding-safe masking and MAP denotes mask-aware attention pooling.
  In the \emph{PSM only} and \emph{None} variants, MAP is disabled and CLS pooling is used for prediction.}
  \label{tab:ablation_integration}
\end{table*}

\begin{figure*}[h]
  \centering

  \begin{minipage}[t]{0.32\linewidth}
    \centering
    \includegraphics[width=\linewidth]{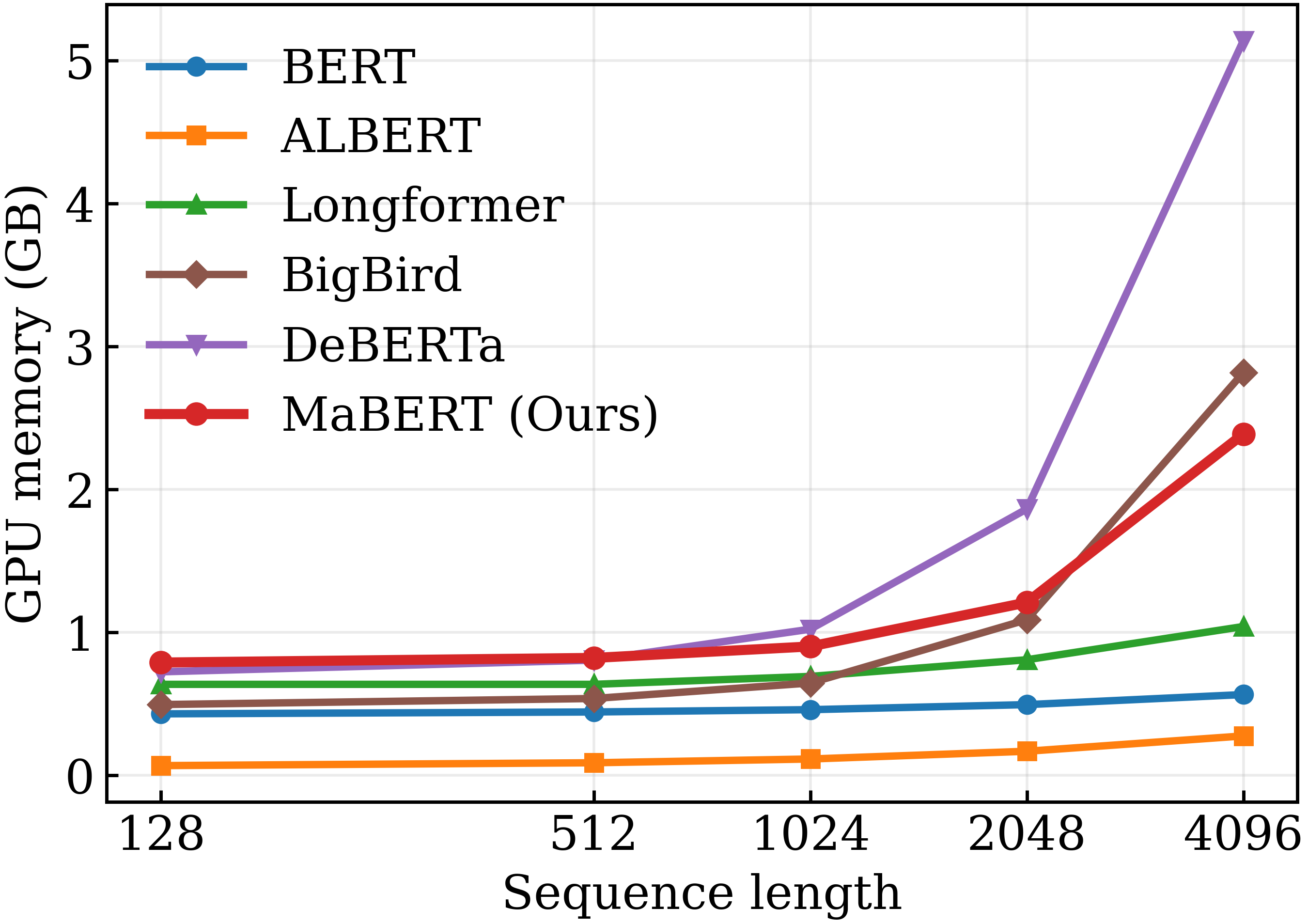}
    \vspace{2pt}
    {\small (a) Peak GPU memory.}
  \end{minipage}\hfill
  \begin{minipage}[t]{0.32\linewidth}
    \centering
    \includegraphics[width=\linewidth]{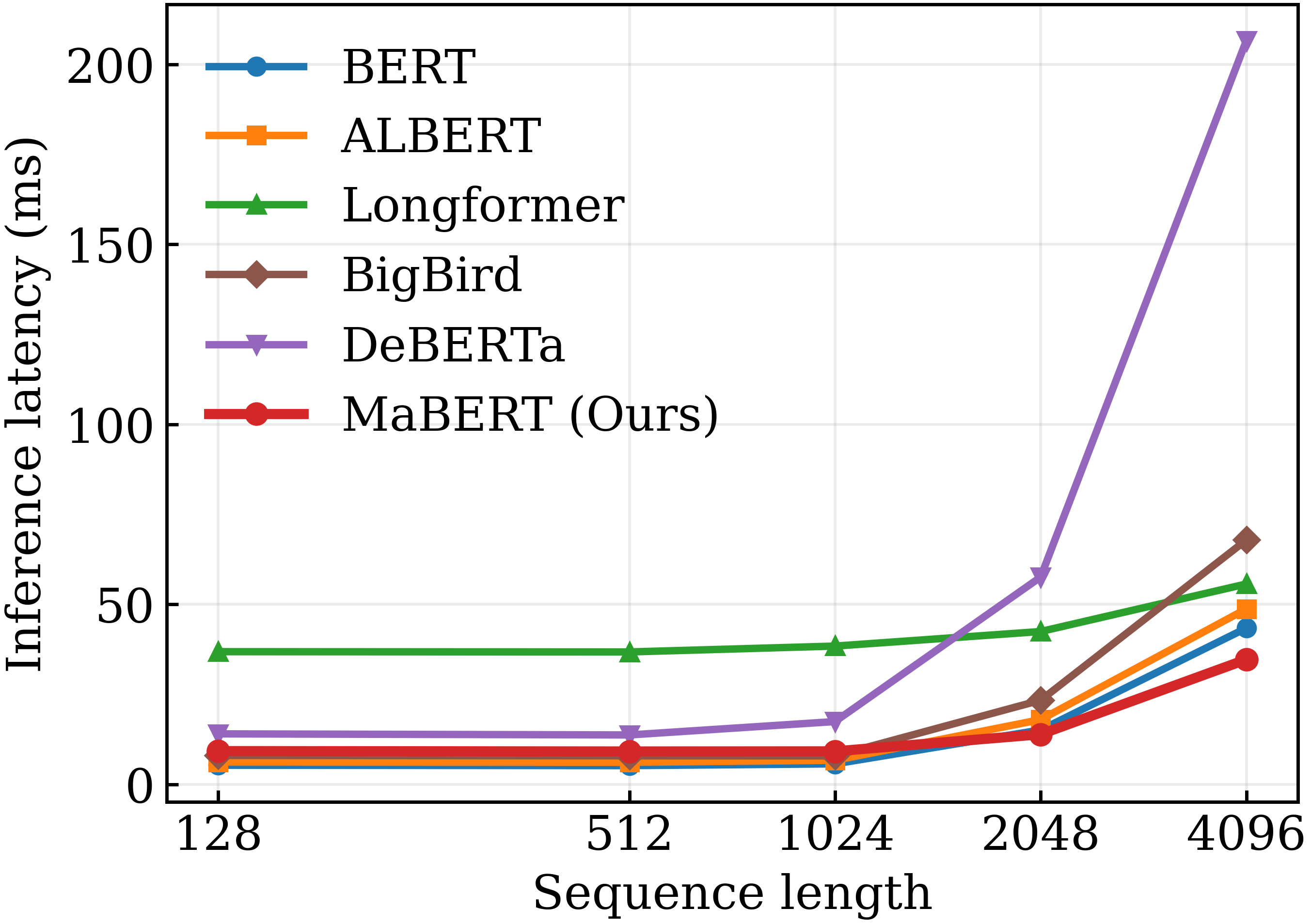}
    \vspace{2pt}
    {\small (b) Inference latency.}
  \end{minipage}\hfill
  \begin{minipage}[t]{0.32\linewidth}
    \centering
    \includegraphics[width=\linewidth]{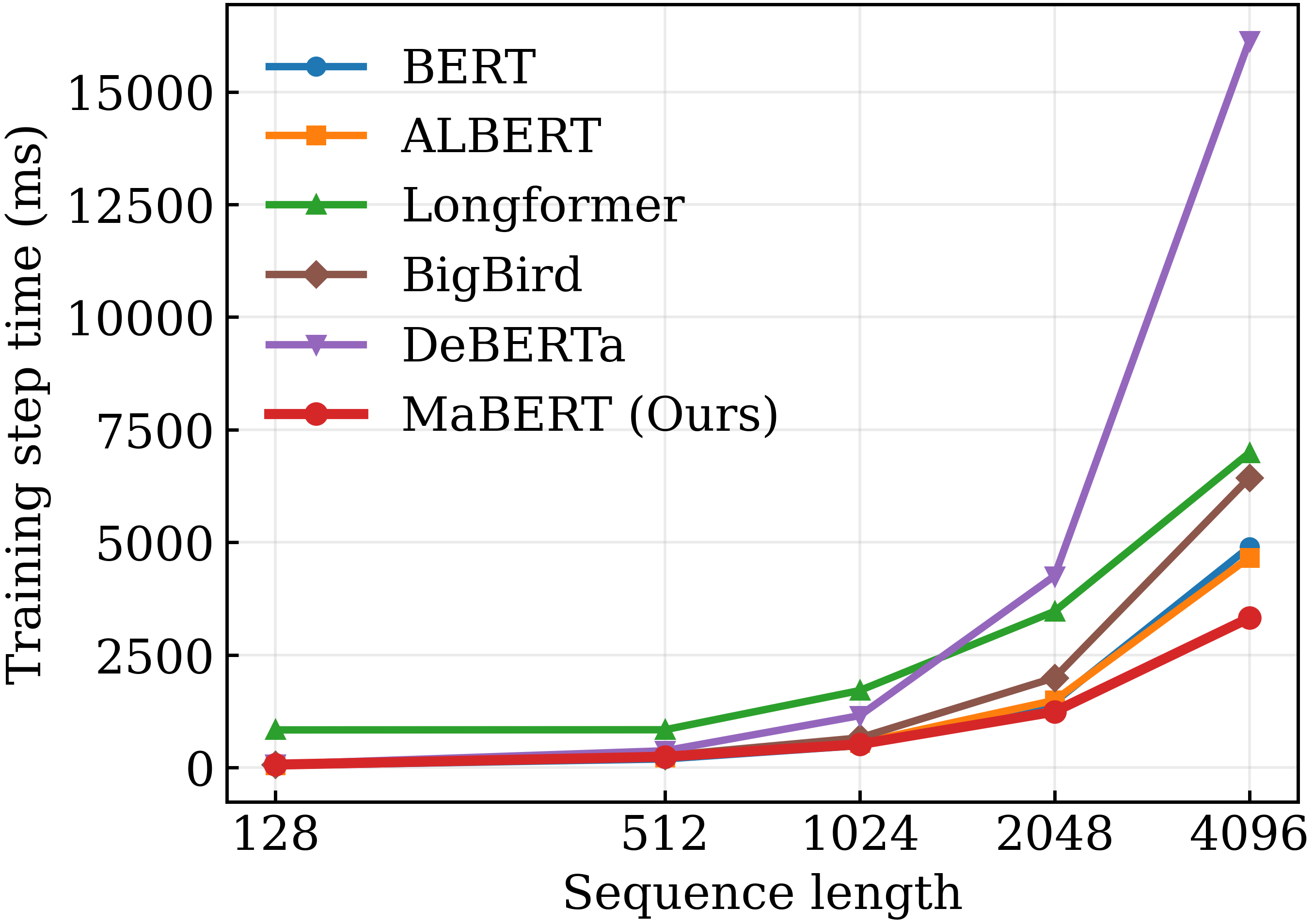}
    \vspace{2pt}
    {\small (c) Training step time.}
  \end{minipage}

  \caption{Efficiency and scalability across sequence lengths.}
  \label{fig:eff_scaling}
\end{figure*}

Table~\ref{tab:ablation_integration} shows that the full model performs best across GLUE. Replacing MAP with [CLS] pooling (“PSM only” and “None”) consistently degrades performance, with the largest drop on CoLA, indicating the importance of valid-token-aware aggregation under variable-length inputs. Removing PSM (“None”) further degrades performance across all tasks, suggesting that suppressing padding-driven accumulation is critical. Overall, PSM and MAP provide complementary gains. Table~\ref{tab:pooling_pred_100pct} further confirms that MAP is the strongest pooling choice across tasks.

Figures~\ref{fig:cola_cosdist_padlen} and~\ref{fig:cola_cosdist_decomp} quantify representation drift as padding length increases with valid tokens held fixed.
We compute cosine distance between padded and unpadded runs using the final-layer [CLS] embedding (Final) or the mean of non-padding token embeddings (unmasked mean).
Without PSM, drift increases with padding; with PSM, drift is strongly suppressed, indicating reduced padding-induced contamination.
The decomposition further shows post-masking outperforms pre-masking, and their combination (Pre+Post) is most stable, highlighting the importance of blocking propagation to upper layers in deep stacks.

\subsection{Efficiency and Scalability}
\label{sec:efficiency}

We compared efficiency and length scalability in terms of peak GPU memory, inference latency, and training step time following Section~\ref{sec:exp_setup}. Peak memory and latency were measured using a single forward pass with a batch size of one (peak memory is the maximum GPU usage during the pass). The training cost was measured as the wall-clock time per optimizer step with an effective batch size of 32. We varied only the input length and evaluated the same checkpoint without additional pre-training.

Figure~\ref{fig:eff_scaling}(a) shows that although MaBERT uses more memory for short inputs, its memory growth is substantially slower with increasing length, resulting in lower peak memory than DeBERTa and BigBird in the long-sequence regime. Figure~\ref{fig:eff_scaling}(b) shows a similar trend for the inference latency: BERT is the fastest for short sequences, whereas MaBERT becomes the most efficient at longer lengths owing to slower latency growth. Figure~\ref{fig:eff_scaling}(c) shows that MaBERT also best mitigates the increase in the training step time as the length increases, whereas DeBERTa exhibits a markedly steeper slowdown. The complete numerical results are reported in Tables~\ref{tab:eff_memory}--\ref{tab:eff_infer} in the Appendix.

\vspace{-0.6em}

\section{Conclusion}\label{sec:conclusion}
We introduced MaBERT, a hybrid MLM-pretrained encoder that interleaves Transformer and Mamba layers, combining global contextual modeling with linear-time state updates while remaining robust to variable-length batching via padding-safe state handling and valid-token aggregation.

With pretraining on BookCorpus and English Wikipedia, MaBERT achieves strong GLUE performance relative to dense and long-context baselines, showing consistent gains on CoLA and sentence-pair tasks. Ablations confirm the value of periodic global interaction, and efficiency results indicate improved memory and runtime scaling with increasing sequence length.

Future work will evaluate MaBERT on long-context understanding and generation benchmarks and study training curricula tailored to extended contexts.

\section{Limitations}
We evaluate MaBERT on GLUE classification benchmarks following MLM pretraining on BookCorpus and English Wikipedia. Although GLUE is a standard testbed for assessing encoder representations, it does not directly measure long-context reasoning, document-level understanding, or generation quality; therefore, our findings primarily reflect sentence- and sentence-pair-level understanding under this protocol. 

In addition, the reported efficiency results are obtained under a fixed hardware and software configuration (e.g., packing and FlashAttention disabled). While we analyze scaling trends across sequence lengths, absolute memory usage and latency may vary depending on optimization strategies, accelerators, and kernel backends.

% =========================
% References
% =========================
\bibliography{custom}
\clearpage

% =========================
% Appendix (A/B sections)
% =========================
\appendix

% Table numbering: A1, A2, ...
\setcounter{table}{0}
\renewcommand{\thetable}{A\arabic{table}}

% ---- (No new packages) helpers ----
% mean ± std in two lines (replaces \makecell)
\newcommand{\mstd}[2]{\begin{tabular}[c]{@{}c@{}}#1\\$\pm$#2\end{tabular}}

% fixed, full-width, NON-float table (inline; caption manual + LEFT-aligned)
\newcommand{\FixedFullWidthTableNF}[3]{%
  \begingroup
  \par\noindent
  \small
  \begin{center}
    \makebox[\textwidth][c]{#1}
  \end{center}
  \vspace{2pt}
  \refstepcounter{table}\label{#2}%
  \begin{minipage}{\textwidth}\raggedright\small
    \textbf{Table \thetable:} #3
  \end{minipage}%
  \par\vspace{14pt}
  \endgroup
}

% ============================================================
% A. Implementation Details
% (keep fixed full-width block to prevent forced page breaks)
% ============================================================
\twocolumn[
\section{Implementation Details}\label{sec:app_impl}
\vspace{0.5\baselineskip}

% =========================
% Table A1: Pretraining recipe
% =========================
\FixedFullWidthTableNF{%
\setlength{\tabcolsep}{6pt}
\renewcommand{\arraystretch}{1.15}
\begin{tabular}{l l}
  \hline
  \textbf{Setting} & \textbf{Value} \\
  \hline
  Data & BookCorpus + English Wikipedia \\
  Objective & MLM only, no NSP \\
  Masking & $p=0.15$, 80/10/10 replacement \\
  Steps (budget) & 1M total steps, 10/25/50/100\% of 1M \\
  Length schedule & 128 for 90\% of steps, 512 for 10\% of steps \\
  Batch size & 256 \\
  Optimizer & Adam, $\beta_1=0.9$, $\beta_2=0.98$, $\epsilon=1\mathrm{e}{-6}$, weight decay $=0.01$ \\
  LR schedule & peak LR $=6\mathrm{e}{-4}$, warmup $=24$k steps, linear decay \\
  Dropout & 0.1 \\
  Tokenizer & Each model's default tokenizer (standard implementation) \\
  \hline
\end{tabular}
}{tab:appendix_recipe}{Common pretraining recipe used for all models.}

% =========================
% Table A2: Efficiency setup
% =========================
\FixedFullWidthTableNF{%
\setlength{\tabcolsep}{6pt}
\renewcommand{\arraystretch}{1.15}
\begin{tabular}{l l}
  \hline
  \textbf{Setting} & \textbf{Value} \\
  \hline
  GPU & NVIDIA A100 80GB (1$\times$GPU) \\
  Precision & bf16 \\
  PyTorch & 2.6.0 \\
  CUDA & 12.2 \\
  Kernel backend & PyTorch SDPA (math/mem-efficient); FlashAttention disabled \\
  Compilation & torch.compile disabled \\
  Packing & disabled \\
  Timing protocol & Median of 100 runs after 20 warmup; torch.cuda.synchronize() before/after timing \\
  \hline
\end{tabular}
}{tab:appendix_eff_setup}{Hardware and software setup for efficiency measurements.}

% =========================
% Table A3: Positional extension
% =========================
\FixedFullWidthTableNF{%
\setlength{\tabcolsep}{6pt}
\renewcommand{\arraystretch}{1.15}
\begin{tabular}{l l}
  \hline
  \textbf{Model} & \textbf{4,096 positional extension} \\
  \hline
  BERT / ALBERT & Resize absolute position embeddings (512 $\rightarrow$ 4,096); initialize new positions via 1D interpolation. \\
  Longformer & Use default long-context setup; set max length to 4,096. \\
  BigBird & Use default long-context setup; set max length to 4,096. \\
  DeBERTa & Extend relative position range to cover 4,096. \\
  MaBERT & Use the same max length and masking rules as in the main text. \\
  \hline
\end{tabular}
}{tab:appendix_pos_ext}{Baseline-specific positional extension for length-4,096 efficiency measurements.}

] % ---- end of fixed full-width block for Section A ----

% -------------------------
% Two-column narrative text for Section A
% -------------------------
\noindent
All models are pretrained under a controlled and unified recipe so that performance differences primarily reflect architectural choices rather than optimization artifacts.
The step-budget protocol enables a compute-matched comparison by truncating training at 10/25/50/100\% of a fixed 1M-step budget while keeping the same length schedule, which isolates the effect of training compute from sequence-length exposure.
Using each model's default tokenizer avoids introducing additional confounds that would not reflect standard usage.

\noindent
Efficiency measurements are conducted under a fixed hardware and software stack with conservative kernel choices to improve repeatability across architectures.
We disable packing, compilation, and FlashAttention, and report median latency over repeated runs with explicit CUDA synchronization, reducing variance from kernel warmup and asynchronous launches.
For length-4,096 runs, baselines are extended with minimal positional-capacity changes appropriate to their design, without altering their core attention or relative-position formulation.

% ============================================================
% B. Additional Results
% (place all tables first using table*, then put text at the end)
% ============================================================
\section{Additional Results}\label{sec:app_results}
\vspace{0.2\baselineskip}
% Table numbering in Section B: B1, B2, ...
\setcounter{table}{0}
\renewcommand{\thetable}{B\arabic{table}}

% ------------------------------------------------------------
% Table A4: GLUE 10%   (table*)
% ------------------------------------------------------------
\begin{table*}[!t]
\centering\small
\setlength{\tabcolsep}{6pt}
\renewcommand{\arraystretch}{1.05}
\begin{tabular}{lcccccccc}
  \hline
  \textbf{Model} &
  \textbf{CoLA} & \textbf{SST-2} & \textbf{MRPC} & \textbf{QQP} &
  \textbf{MNLI-m} & \textbf{MNLI-mm} & \textbf{QNLI} & \textbf{RTE} \\
  \hline
  BERT &
  \mstd{0.419}{0.017} & \mstd{0.874}{0.013} & \mstd{0.821}{0.016} & \mstd{0.837}{0.006} &
  \mstd{0.780}{0.012} & \mstd{0.791}{0.013} & \mstd{0.846}{0.011} & \mstd{0.578}{0.028} \\
  ALBERT &
  \mstd{0.388}{0.019} & \mstd{0.868}{0.013} & \mstd{0.814}{0.017} & \mstd{0.832}{0.007} &
  \mstd{0.772}{0.015} & \mstd{0.781}{0.014} & \mstd{0.842}{0.013} & \mstd{0.566}{0.032} \\
  Longformer &
  \mstd{0.401}{0.020} & \mstd{0.892}{0.014} & \mstd{0.820}{0.016} & \mstd{0.834}{0.008} &
  \mstd{0.776}{0.012} & \mstd{0.787}{0.015} & \mstd{0.858}{0.011} & \mstd{0.573}{0.033} \\
  BigBird &
  \mstd{0.406}{0.019} & \mstd{0.897}{0.012} & \mstd{0.822}{0.016} & \mstd{0.835}{0.005} &
  \mstd{0.781}{0.014} & \mstd{0.792}{0.014} & \mstd{0.862}{0.012} & \mstd{0.577}{0.031} \\
  DeBERTa &
  \mstd{0.423}{0.018} & \mstd{0.881}{0.011} & \mstd{0.817}{0.013} & \mstd{0.829}{0.004} &
  \mstd{0.789}{0.015} & \mstd{0.801}{0.014} & \mstd{0.829}{0.012} & \mstd{0.569}{0.036} \\
  \textbf{MaBERT} &
  \mstd{\textbf{0.574}}{0.016} & \mstd{\textbf{0.904}}{0.009} & \mstd{\textbf{0.837}}{0.016} & \mstd{\textbf{0.868}}{0.003} &
  \mstd{\textbf{0.809}}{0.014} & \mstd{\textbf{0.814}}{0.015} & \mstd{\textbf{0.867}}{0.007} & \mstd{\textbf{0.602}}{0.030} \\
  \hline
\end{tabular}
\caption{GLUE results after pretraining with 10\% of the total steps (mean $\pm$ standard deviation over five seeds; best per task in bold).}
\label{tab:glue_10pct}
\end{table*}

% ------------------------------------------------------------
% Table A5: GLUE 25%   (table*)
% ------------------------------------------------------------
\begin{table*}[!t]
\centering\small
\setlength{\tabcolsep}{6pt}
\renewcommand{\arraystretch}{1.05}
\begin{tabular}{lcccccccc}
  \hline
  \textbf{Model} &
  \textbf{CoLA} & \textbf{SST-2} & \textbf{MRPC} & \textbf{QQP} &
  \textbf{MNLI-m} & \textbf{MNLI-mm} & \textbf{QNLI} & \textbf{RTE} \\
  \hline
  BERT &
  \mstd{0.452}{0.018} & \mstd{0.887}{0.012} & \mstd{0.833}{0.017} & \mstd{0.841}{0.005} &
  \mstd{0.796}{0.011} & \mstd{0.801}{0.021} & \mstd{0.851}{0.010} & \mstd{0.591}{0.027} \\
  ALBERT &
  \mstd{0.438}{0.018} & \mstd{0.884}{0.012} & \mstd{0.828}{0.016} & \mstd{0.840}{0.006} &
  \mstd{0.792}{0.014} & \mstd{0.798}{0.020} & \mstd{0.848}{0.012} & \mstd{0.586}{0.029} \\
  Longformer &
  \mstd{0.455}{0.019} & \mstd{0.906}{0.013} & \mstd{0.836}{0.015} & \mstd{0.844}{0.005} &
  \mstd{0.799}{0.013} & \mstd{0.804}{0.021} & \mstd{0.861}{0.011} & \mstd{0.592}{0.030} \\
  BigBird &
  \mstd{0.462}{0.021} & \mstd{0.910}{0.011} & \mstd{0.839}{0.013} & \mstd{0.846}{0.006} &
  \mstd{0.803}{0.013} & \mstd{0.806}{0.024} & \mstd{0.863}{0.013} & \mstd{0.596}{0.032} \\
  DeBERTa &
  \mstd{0.497}{0.017} & \mstd{0.908}{0.012} & \mstd{0.831}{0.014} & \mstd{0.852}{0.004} &
  \mstd{0.803}{0.014} & \mstd{0.808}{0.019} & \mstd{0.847}{0.011} & \mstd{0.603}{0.034} \\
  \textbf{MaBERT} &
  \mstd{\textbf{0.612}}{0.015} & \mstd{\textbf{0.917}}{0.008} & \mstd{\textbf{0.848}}{0.015} & \mstd{\textbf{0.873}}{0.005} &
  \mstd{\textbf{0.815}}{0.013} & \mstd{\textbf{0.816}}{0.020} & \mstd{\textbf{0.868}}{0.008} & \mstd{\textbf{0.612}}{0.029} \\
  \hline
\end{tabular}
\caption{GLUE results after pretraining with 25\% of the total steps (mean $\pm$ standard deviation over five seeds; best per task in bold).}
\label{tab:glue_25pct}
\end{table*}

% ------------------------------------------------------------
% Table A6: GLUE 50%   (table*)
% ------------------------------------------------------------
\begin{table*}[!t]
\centering\small
\setlength{\tabcolsep}{6pt}
\renewcommand{\arraystretch}{1.05}
\begin{tabular}{lcccccccc}
  \hline
  \textbf{Model} &
  \textbf{CoLA} & \textbf{SST-2} & \textbf{MRPC} & \textbf{QQP} &
  \textbf{MNLI-m} & \textbf{MNLI-mm} & \textbf{QNLI} & \textbf{RTE} \\
  \hline
  BERT &
  \mstd{0.515}{0.018} & \mstd{0.902}{0.010} & \mstd{0.845}{0.016} & \mstd{0.846}{0.005} &
  \mstd{0.807}{0.011} & \mstd{0.813}{0.012} & \mstd{0.859}{0.009} & \mstd{0.598}{0.030} \\
  ALBERT &
  \mstd{0.490}{0.019} & \mstd{0.898}{0.012} & \mstd{0.840}{0.017} & \mstd{0.842}{0.006} &
  \mstd{0.803}{0.014} & \mstd{0.809}{0.013} & \mstd{0.856}{0.011} & \mstd{0.598}{0.031} \\
  Longformer &
  \mstd{0.506}{0.020} & \mstd{0.918}{0.011} & \mstd{0.848}{0.016} & \mstd{0.844}{0.005} &
  \mstd{0.809}{0.013} & \mstd{0.815}{0.014} & \mstd{0.868}{0.010} & \mstd{0.606}{0.033} \\
  BigBird &
  \mstd{0.514}{0.016} & \mstd{0.920}{0.012} & \mstd{0.852}{0.014} & \mstd{0.843}{0.006} &
  \mstd{0.812}{0.012} & \mstd{0.817}{0.018} & \mstd{0.869}{0.012} & \mstd{0.607}{0.030} \\
  DeBERTa &
  \mstd{0.548}{0.018} & \mstd{0.918}{0.011} & \mstd{0.842}{0.015} & \mstd{0.862}{0.004} &
  \mstd{0.811}{0.013} & \mstd{0.820}{0.014} & \mstd{0.868}{0.010} & \mstd{0.619}{0.031} \\
  \textbf{MaBERT} &
  \mstd{\textbf{0.634}}{0.020} & \mstd{\textbf{0.924}}{0.008} & \mstd{\textbf{0.863}}{0.018} & \mstd{\textbf{0.877}}{0.004} &
  \mstd{\textbf{0.825}}{0.011} & \mstd{\textbf{0.828}}{0.021} & \mstd{\textbf{0.886}}{0.007} & \mstd{\textbf{0.624}}{0.037} \\
  \hline
\end{tabular}
\caption{GLUE results after pretraining with 50\% of the total steps (mean $\pm$ standard deviation over five seeds; best per task in bold).}
\label{tab:glue_50pct}
\end{table*}

% ------------------------------------------------------------
% Table A10: Pooling ablation (predicted from 50% ratios)
% ------------------------------------------------------------
\begin{table*}[!t]
\centering\small
\setlength{\tabcolsep}{6pt}
\renewcommand{\arraystretch}{1.05}
\begin{tabular}{lcccccccc}
  \hline
  \textbf{Mode} &
  \textbf{CoLA} & \textbf{SST-2} & \textbf{MRPC} & \textbf{QQP} &
  \textbf{MNLI-m} & \textbf{MNLI-mm} & \textbf{QNLI} & \textbf{RTE} \\
  \hline
  \textbf{MAP} &
  \mstd{\textbf{0.676}}{0.018} & \mstd{\textbf{0.933}}{0.010} & \mstd{\textbf{0.869}}{0.017} & \mstd{\textbf{0.879}}{0.005} &
  \mstd{\textbf{0.835}}{0.016} & \mstd{\textbf{0.837}}{0.017} & \mstd{\textbf{0.893}}{0.012} & \mstd{\textbf{0.654}}{0.033} \\
  ATTN &
  \mstd{0.648}{0.015} & \mstd{0.924}{0.025} & \mstd{0.857}{0.022} & \mstd{0.861}{0.011} &
  \mstd{0.823}{0.011} & \mstd{0.818}{0.023} & \mstd{0.888}{0.016} & \mstd{0.639}{0.036} \\
  CLS &
  \mstd{0.661}{0.011} & \mstd{0.922}{0.031} & \mstd{0.841}{0.023} & \mstd{0.860}{0.004} &
  \mstd{0.819}{0.017} & \mstd{0.820}{0.027} & \mstd{0.878}{0.014} & \mstd{0.597}{0.078} \\
  MaskedMean &
  \mstd{0.651}{0.018} & \mstd{0.924}{0.029} & \mstd{0.857}{0.021} & \mstd{0.866}{0.010} &
  \mstd{0.823}{0.018} & \mstd{0.817}{0.023} & \mstd{0.869}{0.013} & \mstd{0.631}{0.009} \\
  \hline
\end{tabular}
\caption{Effect of pooling choices on GLUE performance for MaBERT (mean $\pm$ standard deviation over five seeds). MAP: mask-aware attention pooling; ATTN: attention pooling; CLS: [CLS]-token pooling; MaskedMean: mean pooling over non-padding tokens.}
\label{tab:pooling_pred_100pct}
\end{table*}

% ------------------------------------------------------------
% Table A7: Memory (table*)
% ------------------------------------------------------------
\begin{table*}[!t]
\centering\small
\setlength{\tabcolsep}{6pt}
\renewcommand{\arraystretch}{1.2}
\begin{tabular}{lrrrrrr}
  \hline
  \textbf{Model} & \textbf{Params} & \textbf{Mem@128} & \textbf{Mem@512} & \textbf{Mem@1024} & \textbf{Mem@2048} & \textbf{Mem@4096} \\
  \hline
  BERT & 109.484 & 441.364 & 454.748 & 471.274 & 507.313 & 579.392 \\
  ALBERT & 11.685 & \textbf{70.649} & \textbf{90.971} & \textbf{117.992} & \textbf{173.382} & \textbf{283.109} \\
  Longformer & 148.661 & 602.629 & 652.634 & 709.375 & 828.660 & 1066.699 \\
  BigBird & 128.061 & 507.797 & 551.183 & 662.196 & 1115.224 & 2885.278 \\
  DeBERTa & 184.417 & 742.110 & 825.994 & 1047.506 & 1908.529 & 5262.076 \\
  MaBERT & 205.322 & 808.787 & 841.846 & 923.875 & 1238.917 & 2445.003 \\
  \hline
\end{tabular}
\caption{Model complexity and peak GPU memory footprint (MB) measured during a forward pass at each sequence length (lowest per column in bold; applied to Mem@* columns).}
\label{tab:eff_memory}
\end{table*}

% ------------------------------------------------------------
% Table A8: Training runtime (table*)
% ------------------------------------------------------------
\begin{table*}[!t]
\centering\small
\setlength{\tabcolsep}{6pt}
\renewcommand{\arraystretch}{1.1}
\begin{tabular}{lrrrrr}
  \hline
  \textbf{Model} & \textbf{Train@128$\times$32} & \textbf{Train@512$\times$32} & \textbf{Train@1024$\times$32} & \textbf{Train@2048$\times$32} & \textbf{Train@4096$\times$32} \\
  \hline
  BERT & \textbf{42.372} & \textbf{189.501} & \textbf{487.204} & 1451.200 & 4894.397 \\
  ALBERT & 50.941 & 225.783 & 545.223 & 1491.872 & 4662.279 \\
  Longformer & 834.994 & 838.428 & 1709.029 & 3466.178 & 6985.296 \\
  BigBird & 60.710 & 263.221 & 658.337 & 1993.710 & 6430.660 \\
  DeBERTa & 74.079 & 371.969 & 1151.794 & 4251.722 & 16136.670 \\
  MaBERT & 64.117 & 240.334 & 516.242 & \textbf{1240.849} & \textbf{3319.429} \\
  \hline
\end{tabular}
\caption{Training runtime (ms) per optimizer step with effective batch size fixed (lowest per column in bold). Positional extension to 4,096 follows Table~\ref{tab:appendix_pos_ext}.}
\label{tab:eff_train}
\end{table*}

% ------------------------------------------------------------
% Table A9: Inference runtime (table*)
% ------------------------------------------------------------
\begin{table*}[!t]
\centering\small
\setlength{\tabcolsep}{6pt}
\renewcommand{\arraystretch}{1.1}
\begin{tabular}{lrrrrr}
  \hline
  \textbf{Model} & \textbf{Infer@128$\times$1} & \textbf{Infer@512$\times$1} & \textbf{Infer@1024$\times$1} & \textbf{Infer@2048$\times$1} & \textbf{Infer@4096$\times$1} \\
  \hline
  BERT & \textbf{5.220} & \textbf{5.331} & \textbf{5.702} & 15.122 & 43.424 \\
  ALBERT & 6.140 & 6.217 & 6.692 & 17.994 & 48.749 \\
  Longformer & 36.794 & 36.866 & 38.420 & 42.468 & 55.676 \\
  BigBird & 7.921 & 7.948 & 8.059 & 23.378 & 67.887 \\
  DeBERTa & 13.737 & 14.038 & 17.456 & 57.592 & 206.592 \\
  MaBERT & 9.160 & 9.193 & 9.263 & \textbf{13.914} & \textbf{34.703} \\
  \hline
\end{tabular}
\caption{Inference latency (ms) per forward pass with batch size 1 (lowest per column in bold). Positional extension to 4,096 follows Table~\ref{tab:appendix_pos_ext}.}
\label{tab:eff_infer}
\end{table*}

% -------------------------
% Narrative text at the end of Section B (after all tables)
% -------------------------

Across compute budgets (Tables~\ref{tab:glue_10pct}--\ref{tab:glue_50pct}), MaBERT consistently outperforms all baselines on every GLUE task.
The largest and most stable gains appear on CoLA across budgets, indicating stronger sensitivity to grammatical acceptability and syntactic regularities under limited and moderate pretraining.
Improvements on MNLI and QNLI persist as compute increases, suggesting that the advantage extends to entailment-focused evaluation settings rather than being confined to single-sentence classification.

Table~\ref{tab:pooling_pred_100pct} compares several pooling strategies for summarizing token-level representations into a single sequence representation for classification.
Overall, mask-aware attention pooling (MAP) yields the strongest and most consistent performance across GLUE tasks, indicating that selectively weighting informative tokens while ignoring padding is beneficial.
Attention pooling (ATTN) and masked mean pooling (MaskedMean) remain competitive on multiple benchmarks but tend to lag behind MAP on tasks that are sensitive to fine-grained sentence properties or entailment cues, such as CoLA, MNLI, and QNLI.
In contrast, CLS pooling shows the largest degradation, most notably on RTE, suggesting that relying on a single terminal representation can be less robust under small-data or high-variance settings.
These results support MAP as the default pooling strategy for MaBERT in downstream evaluation.

Efficiency results (Tables~\ref{tab:eff_memory}--\ref{tab:eff_infer}) highlight distinct scaling behaviors as sequence length grows.
Sparse-attention models reduce memory at moderate long-context lengths, but at 4,096 tokens overhead and implementation details can dominate, leading to less favorable trade-offs than suggested by asymptotic complexity alone.
Compared to dense attention baselines, MaBERT exhibits a substantially lower peak memory footprint at 2,048 and 4,096 than DeBERTa under the same protocol.

Runtime trends mirror the memory observations.
Dense attention remains highly optimized at short contexts, but its cost increases rapidly with length.
At 2,048 and 4,096 tokens, MaBERT achieves the lowest training-step time and the lowest inference latency among compared models, indicating that its long-context efficiency does not rely on sparse attention patterns and translates to practical throughput gains.

\end{document}